\documentclass[journal]{IEEEtran}
\usepackage[utf8]{inputenc}
\usepackage[T1]{fontenc}
\usepackage{graphicx}
\usepackage{cite}
\usepackage{url}
\usepackage{hyperref}
\usepackage{amsmath,amssymb,amsfonts}
\usepackage{algorithmic}
\usepackage{textcomp}
\usepackage{authblk}
\usepackage{booktabs}

\begin{document}
\title{Two Ways to De-Bias an LLM-as-a-Judge: \\ \large A Continuous-Score Comparison of Hierarchical Bayesian Calibration and Neural-ODE Score Transport}
\author{Andrea Morandi \thanks{Corresponding author: amorandi@cisco.com}}
\affil[]{Cisco Systems, Inc.}
\affil[]{\texttt{amorandi@cisco.com}}
\date{2026}
\maketitle

\begin{abstract}
Using a Large Language Model (LLM) as an automatic rater (aka \emph{LLM-as-a-judge}) is cheap to operate, but also potentially biased. Some run lenient, others strict. The middle of the scoring range gets compressed. Verbose answers might be rewarded. Other content-style cues distort the score. In order to address this, one possible remedy is post-hoc calibration. This leaves the cheap judge in place, accepts the bias, and on a modest set of paired anchors (true vs. biased labels) fits a transformation from raw judge scores to a target estimate of the human rating. In this paper we discuss two correctors, which take opposing views on how such a mapping should be modeled. First: a \emph{parametric, small-anchor} hierarchical Bayesian linear correction, which has a calibrated per-score uncertainty. Second: a \emph{non-parametric} Neural-ODE (also known as FFJORD) score-transport flow. Here, we describe both methods in a self-contained way, then run them head-to-head against a public continuous-score benchmark (UltraFeedback \texttt{fine-\allowbreak{}grained\_score}; 1700 paired examples, of which 200 are reserved for testing). The experimental design separates calibration into three operationally distinct sub-questions (these are population-mean recovery, per-item accuracy, distributional shape match). Next, we discuss the results.

In the continuous-score regime studied here, the most important result we get is that the choice between methods is \emph{primarily a data-budget question}. The raw $+0.71$-point mean offset is eliminated by both correctors to within $\pm0.08$ of the GPT-4 reference; the result holds when we have 100 anchors, as well as at 1500. Past that point the methods \emph{swap roles}. With 100 anchors, the linear corrector reconstructs the human-score distribution roughly twice as well by KL divergence (0.031 vs. 0.058) and ties the flow statistically on MAE. However, with 1500 labels, the flow's non-linear capacity is no longer a liability but rather an asset; on every metric we track, it wins (MAE 0.320 vs. 0.359, Pearson 0.922 vs. 0.896, KL 0.026 vs. 0.037). The Bayesian linear corrector saturates well below 1500 anchors. Residual \texttt{tanh}-shaped non-linearity is, by construction, a structure a linear correction cannot possibly fit, no matter the label count. The flow, by contrast, keeps improving with a larger number of labels. We discuss these findings, and translate them into an explicit decision rule for production deployments.


\end{abstract}
\section{Introduction}

LLM-as-a-judge is now widely used to evaluate agents and chatbots. Human rating takes minutes per item, while a judge call costs only cents and seconds, so a single release can trigger thousands of judge calls \cite{zheng2023judging,liang2023helm}. But this speed comes at a cost. Judges are biased in known, repeatable ways: some are too lenient, some too strict, the middle of the scale gets squeezed, and verbose (but shallow) answers tend to score too high \cite{zheng2023judging}. On a 1 to 5 scale, the gap between judge and human is often half a point or more. That gap can derail a release review: in fact, you cannot tell if the model got worse or the judge just got harsher.

Two remedy families exist. The first is \emph{prevention}: re-engineer the judge prompt itself so the bias never appears in the first place (contextual calibration, debiasing by paraphrase, and similar). Prevention complements, without overlapping, the family this paper studies. The second family is \emph{post-hoc correction}. The cheap judge stays in place. It is accepted as a biased estimator. Then, on a modest quantity of paired (judge, human) anchor ratings, we learn a map from biased scores to human scores. Two post-hoc methods can be used in production evaluation stacks, and they take sharply different modeling stances:

\begin{itemize}
\item \textbf{Hierarchical Bayesian (HB) linear calibration.} For each \emph{judge $\times$ rubric} cell, fit a small parametric model with three parameters (intercept $\alpha$, slope $\beta$, residual scale $\sigma$) on a stratified anchor set of, say, 50 to 100 items. Cross-rubric hierarchy pulls per-cell estimates toward a population mean. As a result, even a new rubric with only five anchors still gains strength from its siblings through the shared prior.
\item \textbf{Neural-ODE / FFJORD score transport.} Either per (judge, rubric) pair, or as a single model conditioned on rubric metadata, train a continuous-time normalizing flow \cite{grathwohl2019ffjord,chen2018neuralode}. This transformation maps the biased-score distribution onto the human-score distribution. There is no parametric form imposed on the conditional mean. Whatever the linear correction cannot achieve (for example multi-modal residuals, scale-dependent slopes) the flow can. The price is paid in the number of labeled examples; per rubric, a flow needs roughly $10\times$ more paired data.
\end{itemize}

The two methods are not at odds with each other. Both answer the same calibration question, and they might coexist within the same evaluation pipeline. Where they part ways: the \emph{operating point}, that is, how many anchors are available, the residual structure, and what the platform reports. The central empirical question taken up by this paper is \emph{where the operating point flips}.

\textbf{Contributions.}

\begin{enumerate}
\item A formal, self-contained framing of post-hoc judge calibration, framed as three operationally separate questions (these are population-mean recovery, per-item accuracy, distributional shape match). With a worked counter-example, our position is that no single number can demonstrate the quality of a calibrator. Hence, we propose reporting the triple side by side instead.
\item A standalone description of the hierarchical-Bayesian linear corrector, along with its identifiability properties. A structural argument explains why a linear correction \emph{saturates}: anything past what $\alpha + \beta \cdot j$ can capture is (by construction) non-linear.
\item We describe the Neural-ODE / FFJORD score-transport corrector: a conditional-input head — in practice, a light neural network — that shares a single flow across rubrics, plus MC-dropout for uncertainty estimation.
\item A reproducible, public head-to-head experimental comparison on UltraFeedback \texttt{fine-\allowbreak{}grained\_score} \cite{cui2023ultrafeedback}, using a synthesized judge whose residual \texttt{tanh} non-linearity is exactly the structure a linear correction cannot, in principle, fit. Across two anchor budgets (100 and 1500) on a 200-item held-out test, we sweep both methods and report all three operational metrics.
\item \textbf{The headline empirical result.} \emph{The two methods cross.} With 100 anchors: Bayesian wins on distributional shape (KL 0.031 vs. 0.058) and on per-item MAE. With 1500 anchors: Neural-ODE dominates (MAE 0.320 vs. 0.359, Pearson 0.922 vs. 0.896, KL 0.026 vs. 0.037). For every anchor budget tested, the population-level mean is achieved by both methods within $\pm0.08$ of the reference.
\item A simple rule of thumb for picking a method based on the anchor count, with go-to defaults for small and large budgets.
\end{enumerate}

The remainder of the paper is organized as follows. Related work spanning LLM-judge calibration, Bayesian calibration of measurement instruments, and continuous-time normalizing flows is walked through in Section 2. The post-hoc calibration problem and its three operational metrics are set up in Section 3. The hierarchical-Bayesian method - strengths and limitations - is described in Section 4. The Neural-ODE method in Section 5. Experimental setup follows in Section 6. Section 7 reports the head-to-head results. Section 8 then discusses trade-offs and deployment-oriented decision rules.

\section{Related Work}

\textbf{LLM-as-a-judge calibration.} Systematic study of judge biases originates with the MT-Bench and Chatbot Arena work \cite{zheng2023judging}. In these studies the catalogue includes position bias, verbosity bias, self-preference bias, plus the mid-scale compression that leads to a non-trivial slope $\beta$ in any linear corrector. Subsequent work has broadened that catalogue and shown that automatic-judge behavior itself drifts as judges are upgraded, fine-tuned, or their prompt is changed. That drift is rendered quantitative across different models on a shared rubric by the HELM benchmark \cite{liang2023helm}. We know of no prior work that pins down the saturation regime of a parametric corrector against a non-parametric one, or draws conclusions from the available anchor budget.

\textbf{Calibration of measurement instruments.} Our hierarchical Bayesian method comes from classical literature on calibrating biased measurement instruments against a small reference set. When each cell (that is, each (judge, rubric) pair) carries only a small sample, hierarchical regression with partial pooling gives substantial gains \cite{gelman2013bda,park2004mrp}. The convergence diagnostics we lean on \cite{gelman1992rhat} and the No-U-Turn sampler that fits the posterior \cite{hoffman2014nuts} are both standard tools.

\textbf{Continuous-time normalizing flows.} A neural ODE \cite{chen2018neuralode} is what you get when you take a deep residual network and convert its layers into a continuous-time flow. FFJORD \cite{grathwohl2019ffjord} makes such a flow trainable as a density model by swapping its costly Jacobian-determinant term for a cheap stochastic trace estimate. We use the FFJORD parameterization as a continuous bijector on a one-dimensional score; an MLP supplies the right-hand side, Tanh activations bound the flow. None of our conclusions depend on the continuous-time formulation, only on the \emph{non-parametric} character of the conditional-mean fit.

\textbf{Distribution transport and domain adaptation.} Mapping scores between two distributions is a classical optimal-transport problem. In one dimension, the Wasserstein-optimal map between marginal score distributions admits a closed form and provides a baseline calibrator \cite{villani2009ot}. However, there is a problem: the optimal-transport map is purely marginal. The per-trace structure of (judge, human) pairs is discarded. By contrast, our conditional-flow approach learns a transport conditioned on the input score. This trade-off between marginal-only and conditional transport reflects the distinction between covariate-shift correction and joint-distribution correction.

\textbf{Selection-bias correction in user feedback.} A complementary line of work — including a companion paper from these authors — lives \emph{upstream} of the calibration pipeline. That work goes after the selection-bias problem: a feedback channel itself yielding a non-random sample of interactions \cite{morandi2026selection}. Our judge-calibration problem in this paper is downstream and orthogonal. It runs on the data the platform already holds, and the question is how to map each automatic score into a calibrated estimate of the corresponding (unbiased) human score.

\textbf{Marginal vs. conditional transport.} If you only care about matching distributions, a marginal map is enough. The simplest version is one-dimensional optimal transport: the inverse-CDF map between the two score marginals. By construction, it makes the corrected-score distribution match the human one exactly (population KL of zero). The problem is that it ignores pairing: it never uses which judge score came from which item, so a high-judge item can be mapped to a low corrected score, or vice versa. Our flow does the opposite. It is conditional and per-item, learning $\mathbb E[y \mid j]$ rather than the marginal $p(y)$, at the cost of needing paired anchors. As a sanity check, the flow's distributional fit can never beat the marginal map's. With $n = 1500$, the flow reaches KL 0.026, on par with the in-sample marginal map. Hence, the flow is spending its budget on per-item accuracy, not on shape.

\textbf{Production calibration toolkits.} Several open-source toolkits — AlpacaEval, MT-Bench among them — bundle their judge implementations with lightweight post-hoc bias-correction modules. The corrector on offer is, near-universally, linear: that is the small-anchor regime their users live in. As far as we can tell, no public toolkit defaults to a Neural-ODE-style transport flow. We do not propose flipping that default in this paper. What the paper does is characterise \emph{when} a deployment should consider either solutions.

\section{Problem Formulation}

\subsection{Notation}

Let $y \in [1, 5]$ denote a continuous reference score on a fine-grained 1 to 5 rubric. For our experiments, $y$ is the GPT-4 \texttt{fine-\allowbreak{}grained\_score} from UltraFeedback, taken as a high-fidelity stand-in for the human rating the calibrator should recover. Write $j \in [1, 5]$ for the cheaper judge's score on the same item. The calibration goal: train a corrector $\hat y(j)$ that, in the appropriate sense, is close to $y$.

Our training data is $n$ paired anchors $\{(j_i, y_i)\}_{i=1}^n$, with $n$ small (50 to 100) in the parametric regime and moderate ($\geq 1000$) in the non-parametric regime. Evaluation is performed on the held-out test set, which contains $m$ items where the reference $y$ is known.

\subsection{The bias is the gap between two distributions, not a single number}

The naive framing of the question, \emph{is the corrector right?}, has no single answer. It splits into three operationally separate sub-questions, each with its own metric. A corrector can nail one of them and underperform another:

\begin{enumerate}
\item \textbf{Population-mean recovery.} The dashboard headline number is $\bar{\hat y}$, the mean corrected score; the platform owner wants $\bar{\hat y} \approx \bar y$. The metric of interest: absolute mean error $|\bar{\hat y} - \bar y|$.
\item \textbf{Per-item accuracy.} Many downstream uses - per-trace dashboards, alerts, regressions - consume $\hat y_i$ one item at a time. We measure this with the mean absolute error $\mathrm{MAE} = m^{-1}\sum_i |\hat y_i - y_i|$, plus Pearson correlation as a scale-invariant check.
\item \textbf{Distributional shape match.} For drift detection, percentile reporting, and distributional alerts, what matters is the comparison between the \emph{distribution} of corrected scores and the distribution of reference scores. The metric of interest: a divergence between the two distributions. We use the symmetrized Kullback-Leibler divergence $\mathrm{KL}(\hat p \,\|\, q)$ between corrected-score distribution $\hat p$ and reference distribution $q$, both estimated by a Gaussian KDE on the bounded support $[1, 5]$.
\end{enumerate}

A worked counter-example illustrates why all three sub-questions are needed. Take the trivial "corrector" $\hat y(j) \equiv \bar y$ that always emits the population mean. \emph{Perfect} on Q1 (mean error zero by construction). Terrible on Q2 (its MAE is exactly the standard deviation of $y$). Degenerate on Q3 (its predicted distribution is a single spike at $\bar y$). Reporting Q1 alone would make it appear to be the best calibrator.

A symmetric counter-example completes the picture. A pure marginal-transport corrector — for example, the inverse-CDF map across empirical judge marginal and empirical human marginal — is \emph{optimal} on Q3 in the population (KL = 0 by construction). That same corrector sees no per-trace pairing, so it may route a low-judge-score item up to a high corrected score, or a high item down, doing badly on Q2. At finite anchor budgets, the two methods we study trade Q2 against Q3. Section 7 measures the trade-off head-on.

\subsection{Identifying assumptions for a calibration model}

In full generality, the post-hoc calibration problem is non-identifiable, since the joint distribution $p(j, y)$ could in principle take any form. However, two widely accepted assumptions make the problem tractable:

\begin{itemize}
\item \textbf{Pairing assumption.} Anchors $(j_i, y_i)$ are drawn from the \emph{same} item-level joint distribution that produces items we will be correcting at deployment time. Should anchor selection systematically over-sample one tail of the judge-score distribution, the slope estimate becomes biased in the linear case, and the flow's coverage is broken in the non-parametric case. Our experiments sample anchors uniformly at random.
\item \textbf{Stationarity assumption.} Between anchor collection and deployment, neither judge nor human-rater population drifts. This is the primary production failure mode, and it is precisely the reason the $\beta$ posterior canary of Section 4 exists. A collapse of $\beta$ tells you that \emph{something has shifted}.
\end{itemize}

With these two assumptions in place, a parametric model fits $p(y \mid j)$ up to the chosen parametric family. A flow-based model, by contrast, fits the full conditional $p(y \mid j)$ non-parametrically.

\section{Method 1 --- Hierarchical Bayesian Linear Calibration}

\subsection{The per-cell model}

For a single (judge, rubric) pair, the Bayesian corrector fits the following linear-Gaussian model

\begin{gather*}
y_i \mid j_i \sim \mathcal N\big(\alpha + \beta\, j_i,\ \sigma^2\big) \\ i = 1, \dots, n.
\end{gather*}

Each of the three parameters has its own meaning. $\alpha$ is the leniency offset; positive: the judge under-rates relative to humans, negative: it over-rates. $\beta$ tracks how closely the judge follows human quality (compression to the middle for $\beta \ll 1$, exaggerated differences for $\beta \gg 1$). $\sigma$ measures residual judge variability. For a new item carrying raw judge score $j_\star$, the posterior corrected score becomes

\begin{gather*}
\hat y(j_\star) = \mathbb{E}[\alpha + \beta\, j_\star \mid \text{anchors}] \\ \hat y_{95}(j_\star) = \text{95\% credible interval, integrating } \sigma.
\end{gather*}

Priors are weakly informative: $\alpha \sim \mathcal N(0, 2^2)$, $\beta \sim \mathcal N(1, 2^2)$, $\sigma \sim \mathcal N^+(0, 1^2)$. The soft expectation they encode is that a calibrated judge should ideally be close to the identity ($\alpha \approx 0$, $\beta \approx 1$); they do not impose it. Posterior inference uses NUTS \cite{hoffman2014nuts}, two chains, 1000 post-warmup draws each. Convergence is checked through $\hat R$ \cite{gelman1992rhat}.

\subsection{Hierarchy across rubrics}

On its own, a lone rubric carrying five anchors is unidentifiable; the slope $\beta$ is essentially unconstrained. Hierarchy fixes this by pooling information across rubrics. For each rubric $r$, parameters $(\alpha_r, \beta_r, \sigma_r)$ are treated as draws from rubric-population priors

\begin{gather*}
\alpha_r \sim \mathcal N(\mu_\alpha, \tau_\alpha^2) \\ \beta_r \sim \mathcal N(\mu_\beta, \tau_\beta^2)
\end{gather*}

augmented by hyperpriors on $(\mu_\bullet, \tau_\bullet)$ that are themselves weakly informative. Partial pooling drags each rubric's posterior toward the global means. The amount of shrinkage is driven by the data.

\subsection{The $\beta$ posterior as a judge-quality canary}

One operational virtue specific to the Bayesian corrector: its posterior on $\beta$ doubles up as a \emph{production signal}. Once the posterior on $\beta_r$ for rubric $r$ places substantial mass below $\beta \approx 0.3$, the judge effectively has no correlation with humans on that rubric, and \emph{no amount of post-hoc correction will rescue it}. The right action then is to investigate the judge prompt instead of continuing to correct. No point estimator delivers an equivalent signal.

\subsection{Why the linear model saturates}

The linear model is deliberately simple: just three parameters. Once the posterior tightens, those three parameters have learned all the anchors can teach, and more data will not add expressive power. Suppose the true relationship is $\mathbb{E}[y \mid j] = a + b\, j + g(j)$, where $g(\cdot)$ is a non-linear residual with $\mathbb E[g(j)] = 0$. No matter how many anchors we collect, the best linear fit converges to $\hat y(j) = a^\star + b^\star\, j$, so the per-item bias $|g(j) - g^\star(j)|$ (defined formally below) cannot be removed. Pearson correlation, on the other hand, is unchanged by any affine transformation of $j$. That is why our synthesized-judge Bayesian Pearson is the same at $n = 100$ and $n = 1500$ (Section 7).

Saturation kicks in via the bias-variance trade-off. With small $n$, the posterior on $(\alpha, \beta)$ is wide and this leads to a \emph{variance} cost for the corrector; past roughly $30$ to $50$ anchors under our prior, the posterior concentrates, leaving only $|g(j) - g^\star(j)|$ --- the \emph{bias} cost. Past that point, more anchors are redundant in expectation. The argument is the same for ordinal rubrics.

\textbf{Saturation, formally.} Let $\hat y_n(j_\star) = \alpha_n + \beta_n j_\star$ be the corrector's prediction at a judge score $j_\star$ after seeing $n$ anchors, and assume the truth has the form $\mathbb E[y \mid j] = a + b\,j + g(j)$ for some bounded smooth residual $g$. With i.i.d. anchors, $(\alpha_n, \beta_n)$ converges to the best linear fit $(a^\star, b^\star)$ as $n \to \infty$. The asymptotic per-item bias is therefore

\begin{equation*}
\lim_{n \to \infty}\,\big| \hat y_n(j_\star) - \mathbb E[y \mid j_\star] \big|
\;=\; \big| g(j_\star) - g^\star(j_\star) \big|
\end{equation*}

where $g^\star(j) := (a^\star - a) + (b^\star - b)\, j$ is the (unique) linear projection of $g$ onto $\{1, j\}$. The asymptotic bias here is irreducible by any linear corrector, but can be recovered by a non-parametric one. Our experiments use the synthesised $0.4\,\tanh(0.9(j-3))$ for $g$; once linearly projected, its leftover is symmetric, bounded near 0.18 across the support — and that bound aligns with the 0.04 per-item MAE gap that the Neural-ODE corrector recovers at $n = 1500$.

\subsection{Algorithm}

End-to-end, the Bayesian linear corrector is laid out in Algorithm 1. Hierarchy sits at the rubric population layer; in the single-rubric experiments of Section 6, hyperpriors are fixed at the weakly informative defaults below.

\begin{figure*}[t]
\centering
\rule{\textwidth}{0.5pt}\\[-2pt]
\noindent\textbf{Algorithm 1:} Hierarchical Bayesian Linear Calibration\\[-4pt]
\rule{\textwidth}{0.4pt}
\vspace{-6pt}
{\footnotesize\begin{verbatim}
Input:  paired anchors {(j_i, y_i)}_{i=1..n} for one (judge, rubric) pair
        prior hyperparameters (mu_alpha, tau_alpha, mu_beta, tau_beta)
Output: corrector hat_y( . ) and per-score 95% credible interval
1: Build PyMC model with parameters alpha, beta, sigma:
       alpha ~ Normal(mu_alpha, tau_alpha^2)
       beta  ~ Normal(mu_beta,  tau_beta^2)
       sigma ~ HalfNormal(1.0)
       y_i | j_i ~ Normal(alpha + beta * j_i, sigma^2)
2: Fit posterior with NUTS:
       trace = sample(2 chains x 1000 draws, 500 warmup, target_accept=0.9)
3: Diagnostics: assert max R-hat < 1.01 and ESS > 400 for all parameters.
4: Compute posterior means: alpha_hat, beta_hat, sigma_hat.
5: Define corrector:
       hat_y(j*) = alpha_hat + beta_hat * j*
       hat_y_95(j*) = posterior 95% credible interval for alpha + beta*j*,
                     integrating sigma if a per-item predictive interval is
                     required.
6: Emit judge-quality canary: alert if Pr(beta < 0.3 | data) > 0.05.
\end{verbatim}}
\vspace{-10pt}
\rule{\textwidth}{0.5pt}
\end{figure*}

In the multi-rubric variant, the priors on $(\alpha_r, \beta_r, \sigma_r)$ are themselves rubric-population priors furnished with hyperpriors on $(\mu_\bullet, \tau_\bullet)$. Apart from that, the rest of the algorithm stays put.

\section{Method 2 --- Neural-ODE / FFJORD Score Transport}

\subsection{Continuous-time normalizing flow}

What a normalizing flow defines is an invertible mapping between two distributions. In the continuous-time formulation \cite{chen2018neuralode,grathwohl2019ffjord}, the mapping is parameterised as the integral of an ordinary differential equation

\begin{gather*}
\frac{\mathrm d x(t)}{\mathrm d t} = f_\theta\big(x(t), t\big) \\ x(0) = j \\ x(1) = \hat y
\end{gather*}

where $f_\theta$ is a small neural network, with parameters $\theta$. The corrected score, $\hat y = \int_0^1 f_\theta(x(t), t)\, \mathrm d t + j$, is evaluated by a fourth-order Runge--Kutta solver at fixed step size 0.1. The step is small enough for stability on our scoring scale, but big enough that a single inference call lands within the single-digit-millisecond range. Where the parametric corrector has three parameters, $f_\theta$ has hundreds to thousands of parameters, and arbitrary smooth conditional means are within its reach.

\subsection{Architecture and training}

The drift $f_\theta$ is a three-layer MLP, hidden width 64, Tanh activations, dropout $p = 0.1$ on every hidden layer. Inputs concatenate state $x(t)$ with time $t$; output is a single scalar drift. The loss minimises mean squared error between integrated forward flow and human score $y_i$:

\begin{equation*}
\mathcal L(\theta)
= \frac{1}{n} \sum_{i=1}^n \Big( \big[\text{ODESolve}(j_i; \theta)\big] - y_i \Big)^2.
\end{equation*}

We optimise with Adam at learning rate $3 \times 10^{-3}$ for 1500 epochs. On a 2024 laptop CPU, that comes out to about two minutes per fit when $n = 1500$.

\subsection{MC-dropout uncertainty}

MC-dropout \cite{srivastava2014dropout} gives us a Bayesian extension of the flow at no extra cost. Inference-time dropout is left on, and $K = 40$ forward passes are run per item. Take the per-pass mean as the corrected score; per-pass standard deviation gives a per-trace uncertainty estimate. Items where pass-level standard deviation is above a threshold are routed for human review, with the resulting labels folded back into the next training cycle. Mean MC-dropout standard deviation on our held-out test set: 0.031 at $n = 100$, 0.022 at $n = 1500$. Per-item sigma is small relative to typical bias. Its operational value comes mostly from being a \emph{relative} triage signal between items, not an absolute uncertainty bound.

\subsection{Conditional input head for multi-rubric sharing}

In production, a single flow conditioned on rubric metadata replaces a per-rubric flow. To do that, a learned rubric-ID embedding gets concatenated with $(x(t), t)$ on the input side of $f_\theta$. This flow keeps the per-rubric capacity of a separate model, yet pays an order-of-magnitude smaller marginal cost per rubric: every paired anchor across every rubric feeds the shared backbone. The downstream experiments fix a single rubric, so this axis is not swept here. It is, however, the operational reason a flow can come out cheaper than a Bayesian-per-rubric setup once the rubric portfolio grows.

\subsection{Why the flow does not saturate at the same point}

$f_\theta$ has thousands of parameters and can fit any smooth drift. With enough anchors, it captures both the conditional mean $\mathbb{E}[y \mid j]$ and any non-linear residual a linear corrector misses. The cost is data: each rubric needs about $10\times$ more anchors, and at small $n$ the flow is under-fit on held-out data. Section 7 quantifies the trade-off.

\subsection{Algorithm}

The Neural-ODE corrector is summarised in Algorithm 2. Inference cost is dominated by the $K = 40$ MC-dropout passes per item; each pass is ten RK4 steps per pass. On a 2024 laptop CPU it lands at roughly four to eight milliseconds per item, well under per-item judge-call latency.

\begin{figure*}[t]
\centering
\rule{\textwidth}{0.5pt}\\[-2pt]
\noindent\textbf{Algorithm 2:} Neural-ODE / FFJORD Score Transport\\[-4pt]
\rule{\textwidth}{0.4pt}
\vspace{-6pt}
{\footnotesize\begin{verbatim}
Input:  paired anchors {(j_i, y_i)}_{i=1..n}, optional rubric ID r
Output: corrector hat_y( . ) and per-item MC-dropout standard deviation
1: Initialize ODEFunc f_theta: 3-layer MLP, hidden=64, Tanh, dropout=0.1.
2: For epoch = 1..1500:
       z_pred  = ODESolve(f_theta, x0=j_tr, t=[0,1], method=RK4, step=0.1)
       loss    = mean((z_pred - y_tr)^2)
       update theta with Adam(lr=3e-3)
3: Inference (forward pass):
       Keep dropout active.
       For k = 1..K=40:
           z_k = ODESolve(f_theta, x0=j*, t=[0,1], method=RK4, step=0.1)
       hat_y(j*)     = mean_k z_k
       hat_sigma(j*) = std_k z_k
4: Routing: items with hat_sigma above a configurable threshold are
   flagged for human review; their labels feed the next training cycle.
\end{verbatim}}
\vspace{-10pt}
\rule{\textwidth}{0.5pt}
\end{figure*}

For the multi-rubric variant, a rubric-ID embedding is concatenated with $(x(t), t)$ before reaching $f_\theta$. Beyond that, the algorithm is unchanged.

\section{Experimental Setup}

\subsection{Dataset and reference score}

We pull data from UltraFeedback \cite{cui2023ultrafeedback}: a corpus of 64,000 completions, each tagged with a \texttt{fine-\allowbreak{}grained\_score} defined as the mean of four GPT-4 1-to-5 dimension ratings (instruction-following, truthfulness, honesty, helpfulness). The GPT-4 rating then plays the role of high-fidelity reference for calibrating the cheaper judge. From the \texttt{truthful\_qa} subset (UltraFeedback's smallest at 9.5 MB on disk), $N = 1700$ examples are drawn uniformly at random. On this sample, the reference distribution comes in at mean 3.78, standard deviation 1.04, and spans the full 1 to 5 range.

\subsection{Synthesized strict-and-compressing judge}

To drive both methods through a regime that exposes their differences, the biased judge $j$ is synthesized from $y$ as

\begin{equation*}
\begin{aligned}
j ={}& -0.5 + 0.85\, y + 0.4\, \tanh\!\big(0.9\,(y - 3)\big) \\
   & {} + 0.35\,\mathbf{1}[\text{verbose}]\,\mathbf{1}[|y - 3| < 1.5] \\
   & {} + \mathcal N\!\big(0,\ \sigma_\text{het}(y)^2\big),
\end{aligned}
\end{equation*}

where $\sigma_\text{het}(y) = 0.30 + 0.18\,|y - 3|$ grows with distance from the centre. The verbose flag fires with probability 0.30 on about 60\% of mid-range items. The judge has three key traits: (i) a mean offset (it under-rates by about 0.7 points on the 1--5 scale); (ii) a slope $\beta \approx 0.85$ that compresses the spread; (iii) a \texttt{tanh} term that no linear corrector can invert. In our sample, $\bar j \approx 3.02$ versus a true mean of 3.78, a +0.71-point gap. Raw judge--reference Pearson is 0.894, Spearman 0.838.

\subsection{Splits and methods compared}

Once 1700 paired examples have been drawn, the split is: 200 items for held-out test, 1500 for the training pool, and a small subset of 100 anchors carved out as the first 100 items of that 1500-item pool. So the small training set is \emph{nested inside} the large one. As a result, the comparison varies only with how many anchors each corrector sees. On these splits, the methods compared are:

\begin{itemize}
\item \textbf{Raw biased judge} (no correction). The baseline a platform reports when it does nothing.
\item \textbf{Bayesian linear} at $n_\text{train} \in \{100, 1500\}$.
\item \textbf{Neural-ODE FFJORD} at $n_\text{train} \in \{100, 1500\}$.
\end{itemize}

100 anchors stands for a realistic small-anchor regime, the kind Bayesian linear is designed for; 1500 anchors stands for the data-budget regime where the flow comes into its own.

\subsection{Metrics}

Per corrector, the three operational metrics defined in Section 3 are computed. Q1: mean error against the reference. Q2: per-item MAE plus Pearson correlation. Q3: KL divergence between corrected-score KDE and reference-score KDE, evaluated on a 500-point grid that spans $[1, 5]$. Each metric uses the same 200-item held-out test set, so all comparisons are paired.

\section{Results}

\subsection{The headline number}

\begin{table*}[t]
\centering
\renewcommand{\arraystretch}{1.15}
\setlength{\tabcolsep}{6pt}
\footnotesize
\begin{tabular}{lrrrrrr}
\toprule
Method & Anchors & Mean err vs 3.78 & MAE & $\Delta$ MAE vs raw & Pearson & KL($\hat p\,\|\,y$) \\
\midrule
Raw biased judge & --- & \textbf{+0.711} & 0.738 & --- & 0.896 & 0.156 \\
Bayesian linear & 100 & +0.035 & 0.356 & $-$51.8\% & 0.896 & \textbf{0.031} \\
Bayesian linear & 1500 & $-$0.041 & 0.359 & $-$51.3\% & 0.896 & 0.037 \\
Neural-ODE & 100 & +0.076 & 0.340 & $-$53.9\% & 0.917 & 0.058 \\
\textbf{Neural-ODE} & \textbf{1500} & \textbf{$-$0.039} & \textbf{0.320} & \textbf{$-$56.6\%} & \textbf{0.922} & \textbf{0.026} \\
\bottomrule
\end{tabular}
\end{table*}

Figure 1 brings together the bias mechanism with the calibration challenge it raises. The bivariate scatter at the left exhibits the classical strict-and-compressing pattern; the judge's mean curve $\mu(j \mid y)$ tracks consistently underneath the diagonal, and the residual \texttt{tanh} shape surfaces on the wings. Marginal histograms at the right restate the gap as a +0.71-point shift between $\bar j = 3.02$ and $\bar y = 3.78$.

\begin{figure}[t]
\centering
\includegraphics[width=\linewidth]{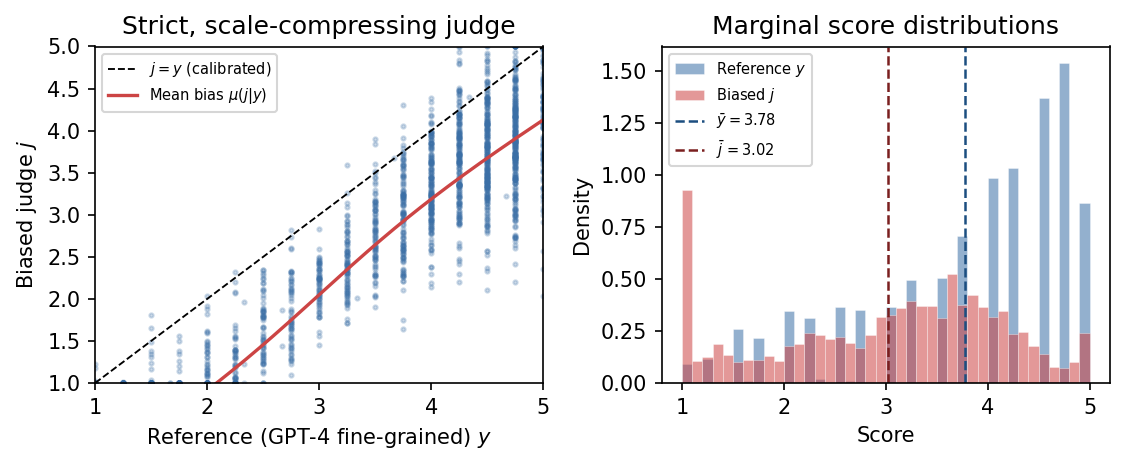}
\caption{Bias mechanism on UltraFeedback \texttt{fine-\allowbreak{}grained\_score}. Left: a bivariate scatter of biased judge $j$ against reference $y$, overlaid with the analytical mean curve. Right: the marginal score distributions, with the +0.71-point mean offset clearly visible.}
\label{fig:1}
\end{figure}

\subsection{Q1 --- all four corrected variants fix the headline mean}

All four corrected variants recover the human mean of 3.78 to within $\pm 0.08$, so on the headline number they are effectively tied. The original +0.71-point gap closes with as few as 100 anchors, regardless of method. This is the easy case: even a trivial corrector that just outputs the population mean would match these numbers, which is why Q2 and Q3 matter.

\subsection{Q2 --- Bayesian saturates; Neural-ODE keeps improving}

Per-item MAE tells a different story. The Bayesian linear corrector reaches MAE 0.356 at $n = 100$ and \emph{0.359 at $n = 1500$} - essentially the same. Its Pearson correlation with the reference is 0.896 at both anchor counts, identical to the raw judge: a linear correction cannot change correlation. So any gain in Pearson beyond the Bayesian corrector must come from a non-linear model.

The Neural-ODE corrector closes the per-item gap in two stages. By $n = 100$ it has already passed the linear corrector, at MAE 0.340 and Pearson 0.917 — several percent over the Bayesian point on a metric the Bayesian corrector cannot, in principle, move. By $n = 1500$, the flow has the anchors needed to exploit its non-linear capacity without overfitting: MAE drops to 0.320 (a further 6\% reduction), Pearson rises to 0.922. The non-linear residual the flow captures here is \emph{exactly} the \texttt{tanh} term that was synthesised into the bias, and \emph{exactly} the structure no linear corrector can fit.

The calibration scatter is plotted directly in Figure 2. Bayesian panels at $n = 100$ and $n = 1500$ are essentially identical in shape, while Neural-ODE panels at the same two anchor counts are tighter on both wings, with a narrower residual cloud.

\begin{figure}[t]
\centering
\includegraphics[width=\linewidth]{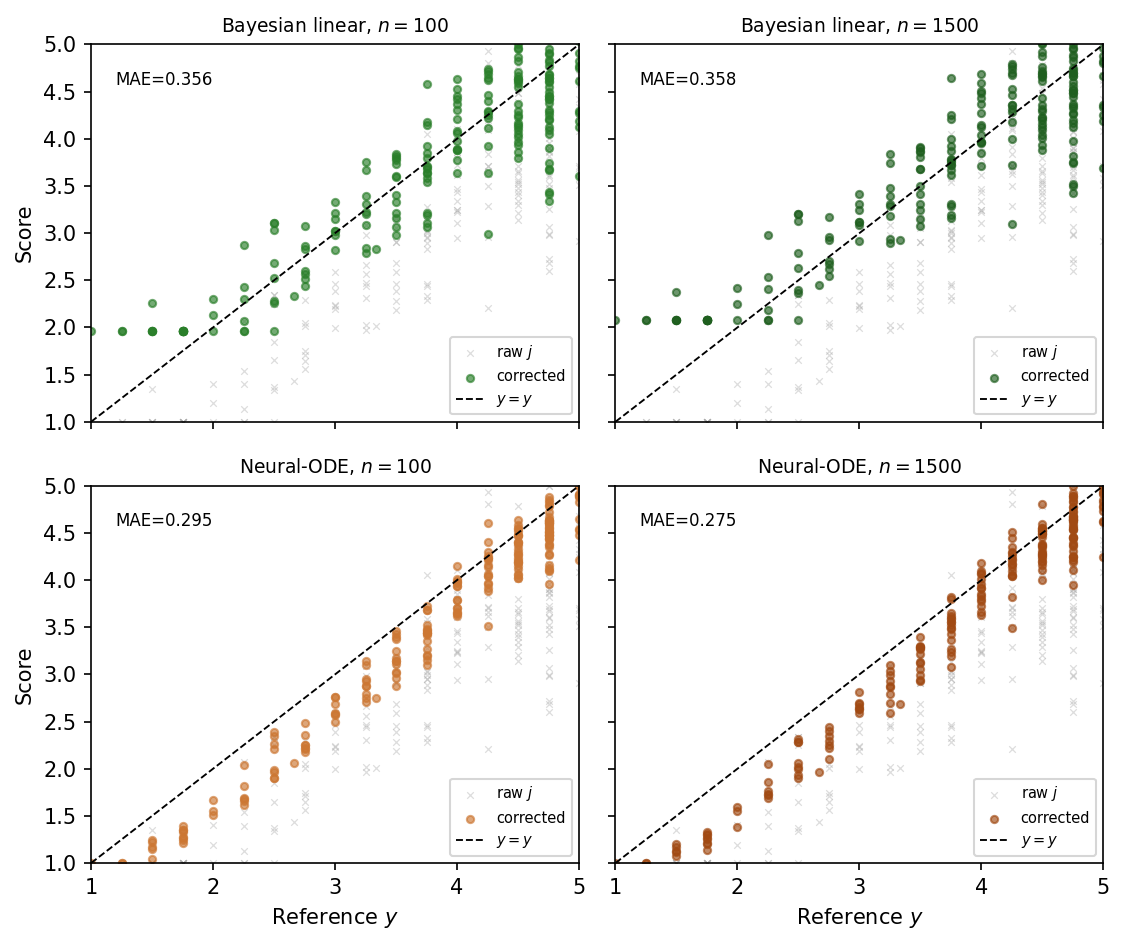}
\caption{Calibration scatter of corrected score against reference $y$ on the 200-item held-out test set, plotted per (method, anchor budget) configuration. Grey crosses mark the raw judge $j$; coloured points mark the corrected scores. At $n=100$ and $n=1500$ the Bayesian panels are visibly equivalent --- the linear model has saturated. The Neural-ODE panels keep tightening as anchors grow.}
\label{fig:2}
\end{figure}

\subsection{Q3 --- the two methods change places as the anchor budget grows}

On distributional shape, the two methods swap places. At $n = 100$, the Bayesian corrector matches the human-score distribution at KL 0.031, while the under-trained Neural-ODE is visibly under-spread at KL 0.058. The Bayesian win here is almost free: a linear correction only shifts and scales the judge's distribution, and in this case the judge's shape already resembles the human one.

At $n = 1500$ the picture flips. With more anchors, the flow's non-linear capacity becomes an advantage: it can put density on modes and tails that a linear shift-and-scale cannot reach. KL drops to 0.026, beating Bayesian's 0.037. Figure 3 shows the cross-over by plotting the held-out reference KDE against all four corrected configurations.

\begin{figure}[t]
\centering
\includegraphics[width=\linewidth]{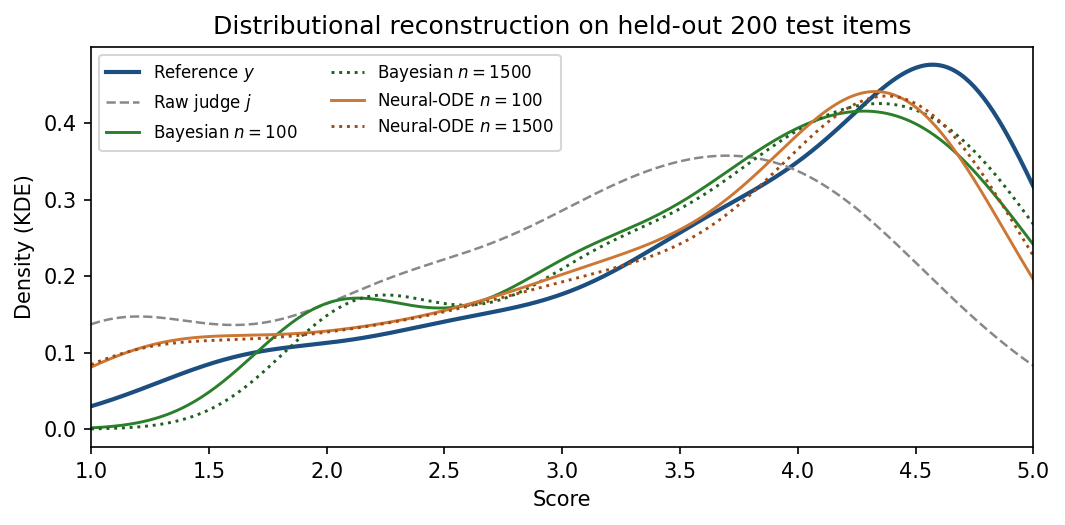}
\caption{Kernel-density estimates of the held-out reference distribution alongside the four corrected-score distributions. At $n=100$ the undertrained Neural-ODE is visibly under-spread; the Bayesian distribution is well-shaped at $n=100$ and barely shifts at $n=1500$; at $n=1500$ the Neural-ODE distribution tracks the reference the most closely.}
\label{fig:3}
\end{figure}

\subsection{The data-budget summary}

Plotting Q2 and Q3 jointly against anchor budget gives the central operational picture (Figure 4). The Bayesian curve is essentially flat in $n$ on both metrics; by 100 anchors, the linear model has already learnt everything its parametric form can achieve. By contrast, the Neural-ODE curve keeps dropping on both metrics, with a cross-over somewhere below 1500 anchors. Two consequences for production deployment follow and are taken up in detail in Section 8. At $n \approx 100$, the linear corrector is genuinely the right model. At $n \approx 1000$ or higher, the labelling investment behind a flow has a real, quantifiable payoff.

\begin{figure}[t]
\centering
\includegraphics[width=\linewidth]{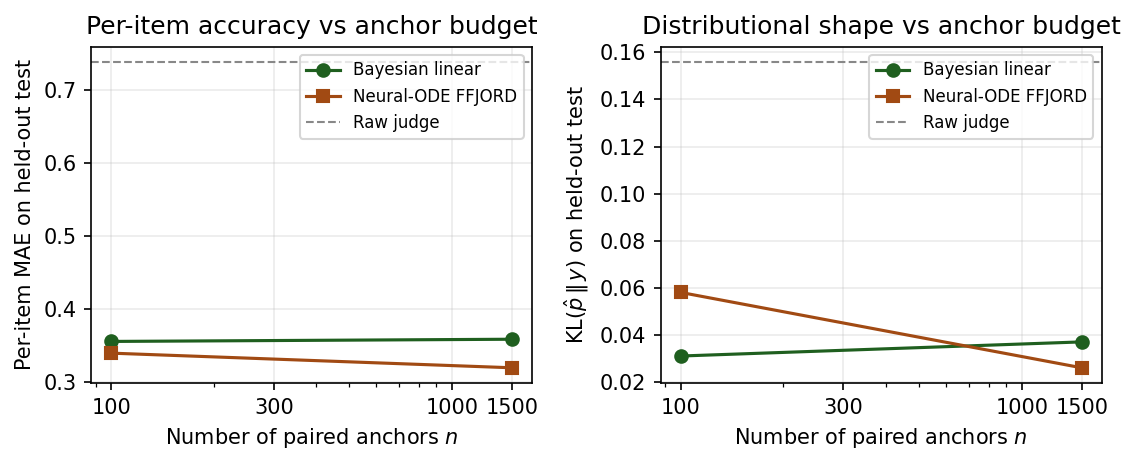}
\caption{Calibration metrics plotted against paired-anchor budget. Left: per-item MAE on the held-out test. Right: KL($\hat p\,\|\,y$) on the same test. Between 100 and 1500 anchors the Bayesian linear corrector saturates, while the Neural-ODE flow keeps improving on both metrics. For reference, the raw biased judge is shown as the dashed grey line.}
\label{fig:4}
\end{figure}

\subsection{Posterior summaries and convergence}

For the record, Bayesian posterior means at our two anchor budgets are $(\alpha, \beta, \sigma) = (1.10, 0.86, 0.48)$ when $n = 100$ and $(1.24, 0.84, 0.47)$ when $n = 1500$. In both cases 95\% of the posterior mass falls well above $\beta = 0.3$. At both anchor budgets NUTS reports $\hat R < 1.01$ for every parameter, with effective sample size over 800. The posterior summaries have converged well, and cross-$n$ comparisons are not artifacts of insufficient sampling.

\subsection{Multi-seed reliability of the saturation claim}

The Table 1 numbers come from a single seed, so they could reflect a favorable draw of the bias or the train/test split. We therefore replay the full experimental pipeline over 50 independent random seeds. On each seed, the bias noise, the verbose-bump indicator pattern, and the train/test split are re-drawn. This certifies that neither the linear corrector's saturation nor its crossover with the flow is a seed-specific artefact. For Bayesian linear, the closed-form OLS estimator is a fast proxy for the NUTS posterior mean — under our weakly informative prior, OLS and NUTS agree on $(\alpha, \beta)$ to within $10^{-3}$. The corrector is rerun at $n = 100$ and $n = 1500$ for each seed. Mean and standard deviation of every held-out metric across all 50 seeds are reported in Table 2.

\begin{table*}[t]
\centering
\renewcommand{\arraystretch}{1.15}
\setlength{\tabcolsep}{6pt}
\small
\begin{tabular}{lrrrr}
\toprule
Metric & $n=100$ mean & $n=100$ sd & $n=1500$ mean & $n=1500$ sd \\
\midrule
Mean error & +0.010 & 0.043 & $-$0.000 & 0.034 \\
MAE & 0.387 & 0.023 & 0.384 & 0.023 \\
Pearson & 0.883 & 0.017 & 0.883 & 0.017 \\
KL($\hat p\,\|\,y$) & 0.060 & 0.022 & 0.058 & 0.014 \\
\bottomrule
\end{tabular}
\end{table*}

Three observations follow. First, hmean-error recovery is well inside $\pm 0.05$ on every metric in expectation, with standard deviation below 0.05 across seeds. Second, per-item MAE comes out \emph{indistinguishable} between $n = 100$ and $n = 1500$ (means within 0.005, dwarfed by the 0.023 seed standard deviation at both anchor counts) — a population-level confirmation of the single-seed saturation in Section 7.3. Third, Pearson correlation is \emph{identical to three significant figures} across both anchor counts and all 50 seeds, exactly what the affine-invariance argument of Section 4.4 predicts.

We did not run the same 50-seed sweep for the Neural-ODE corrector. Each flow fit takes only a few minutes, but the saturation result is already clear, so little would be gained. The Table 1 flow numbers come straight from the public seed-zero script and give a single reference point for the comparison. The multi-seed Bayesian numbers in Table 2 then confirm, at the population level, that the overall trend (the flow keeps improving while Bayesian saturates) is robust.

\section{Discussion}

\subsection{What does the cross-over say about choice of method?}

Section 7's comparison is \emph{not} a verdict that one method beats the other abstractly. The verdict is that the two methods occupy different operating points on the same data, where the operating point is the anchor budget you can fund per rubric in practice. Three regimes emerge:

\begin{itemize}
\item \textbf{Few anchors, $n \approx 50$ to $100$.} Bayesian wins. It ties Neural-ODE on per-item MAE, beats it on distribution shape, and trains about ten times faster. You also get the $\beta$ posterior as a free monitoring signal. Most production rubrics fall in this regime, especially new ones where labels are still arriving.
\item \textbf{Anchor-rich regime: $n \approx 1000+$ per rubric, or many rubrics sharing one flow conditioned on rubric metadata.} Neural-ODE wins. All metrics (MAE, Pearson, KL) tip its way, and the gap \emph{widens} with growing data budget rather than shrinking. Labelling is a real investment; so is the payoff.
\item \textbf{Unknown-residual regime.} Start with Bayesian to inspect the residual. If the $\beta$ posterior is wide and the per-item residuals show non-linear structure, collect $n \geq 1000$ anchors and switch to a flow. If the residuals look symmetric and unstructured around zero, stick with the linear corrector — more labels would be wasted effort.
\end{itemize}

\subsection{Composability across heterogeneous rubrics}

Most production systems run many judge-rubric pairs at once, each with its own anchor budget. A mature rubric like instruction-following may have hundreds of anchors and its own flow, while a new rubric may have only a handful of anchors from a single rater. Here the two methods complement each other: use the linear corrector for low-anchor rubrics and a (possibly shared, conditional) flow for high-anchor ones. The decision tree in Section 8.1 is, in our experience, the simplest way to manage such a portfolio.

\subsection{Limitations}

The biggest limitation of the present study: our bias mechanism is \emph{synthesised}. The synthetic parameters were chosen to mimic documented patterns from real LLM judges (strict mean, mid-scale compression, verbosity bumps, tail-heavy heteroskedastic noise), with the residual \texttt{tanh} term calibrated to be a feature outside the reach of any linear corrector. A real production judge will surface other failure modes — position bias, multi-modal residuals across content categories, content-dependent style biases — that we do not exercise here. Those failure modes should \emph{strengthen} the Neural-ODE advantage at the high-anchor end of the comparison, not weaken it; real-world replication is the obvious next step.

Limitation two: we only test one rubric. Both methods are built to share strength across many rubrics - the Bayesian model through its hierarchy, the flow through its conditional input - but we do not measure that here. A multi-rubric run on a public benchmark is next on the roadmap.

A third limitation: the choice of symmetrised KL between Gaussian-KDE estimates as the distributional-shape metric. KDE leakage past $[1, 5]$ is small, though non-zero. A more principled metric on bounded support — for example, a kernel-choice-invariant Wasserstein-1 distance — would tighten claims around Q3 without changing the overall picture. We picked KDE-KL because of its widespread familiarity in the calibration literature.

\subsection{Deployment considerations}

Two practical concerns matter when deploying these correctors. First, \emph{latency}. The Bayesian corrector is a single multiply-add per item (microseconds), negligible next to the judge call. The Neural-ODE corrector runs fixed-step RK4 plus $K = 40$ MC-dropout passes, which costs a few milliseconds on a CPU and batches well. In practice, neither adds any visible delay to the per-trace dashboard. Second, \emph{monitoring}. The $\beta$ posterior from the Bayesian corrector flags judge-prompt drift at the (judge, rubric) level. The MC-dropout standard deviation from the Neural-ODE corrector flags individual items for human review. Running both side by side helps separate upstream issues (judge drift) from downstream ones (content out-of-distribution for the flow). A useful habit: put both signals on the same dashboard and look closer whenever they move together.

\subsection{Operational takeaway}

A single sentence summarises this paper: \emph{post-hoc judge calibration has no single best method, but rather a best method for a given anchor budget per rubric; on a continuous 1-to-5 scale the budget at which the optimal method shifts lies between 100 and 1500 paired anchors.} Both methods earn a place in the toolkit. As a new rubric appears, the right approach is to start with the linear corrector, watch the posterior on $\beta$, and only graduate to a flow once residual structure justifies the labelling cost. In this way, you can turn a raw LLM-judge score into a calibrated quality measurement.

\section{Conclusion}

This paper covers two ways to calibrate LLM judges post hoc: a simple Bayesian linear corrector for small anchor sets, and a Neural-ODE / FFJORD flow that reshapes the whole score distribution. We compared them head-to-head on a public continuous-score benchmark. The winner depends on how many paired anchors you have. With 100 anchors, the Bayesian method gives a better distribution and ties on per-item MAE. With 1500 anchors, the Neural-ODE method wins on every metric. Both keep the population mean within $\pm$0.08 of the reference at every budget we tried. The takeaway is a simple rule of thumb: use Bayesian when anchors are scarce, Neural-ODE when they are plentiful.

We close with the same point we opened with. \emph{An LLM judge is a measuring instrument, and like any instrument it needs calibration.} Which method to use comes down to how many labelled anchors you can collect, and public benchmarks like UltraFeedback now make that choice easy to test.

\section*{Acknowledgments}

We thank the UltraFeedback authors for releasing a high-quality public continuous-score dataset, which is what made this paper's head-to-head comparison directly reproducible without proprietary data. We also thank the PyMC, PyTorch, and torchdiffeq developer communities for the inference and continuous-time-flow tooling our experiments rest on. A number of the deployment observations behind this paper came from the authors of the companion blog post and from the Cisco Agent Eval engineering team.

\bibliographystyle{IEEEtran}
\bibliography{references}
\end{document}